\documentclass[letterpaper, 10 pt, conference]{ieeeconf}  %

\IEEEoverridecommandlockouts                  
\usepackage{cancel}
\usepackage{xcolor}

\usepackage{amsmath}
\usepackage{amssymb}  %
\usepackage{mathtools}
\usepackage{transparent}

\usepackage{import}
\usepackage{pgfplots}
\pgfplotsset{compat=1.18}
\usepackage{color}
\usepackage{tabularx} %
\usepackage{multirow}
\usepackage{pgf}
\usepackage{blindtext}
\usepackage{siunitx}
\usepackage{balance}
\usepackage{epsfig}
\usepackage{comment}
\usepackage{textcomp}
\usepackage{gensymb}
\usepackage{xfrac}

\newcommand{\vect}[1]{\boldsymbol{#1}}
\newcommand{\matr}[1]{\boldsymbol{#1}}

\newcommand{\Ap}{\ensuremath{\frac{\partial \vect{f}_p}{\partial \vect{x}}}}
\newcommand{\ApT}{\ensuremath{\left(\frac{\partial \vect{f}_p}{\partial \vect{x}}\right)^\mathsf{\!\!\!T}}}

\newcommand{\iBp}{\ensuremath{{\partial\vect{f}_p}/{\partial\vect{u}}}}
\newcommand{\BpT}{\ensuremath{\left(\frac{\partial \vect{f}_p}{\partial \vect{u}}\right)^\mathsf{\!\!\!T}}}

\newcommand{\dotxik}[1]{\ensuremath{\dot{\vect{\xi}}{\vphantom{\vect{\xi}}}^{#1}}}
\newcommand{\dotthetak}[1]{\ensuremath{\dot{\vect{\theta}}{\vphantom{\vect{\theta}}}^{#1}}}

\newcommand{\IEEEacceptance}{
    \begin{tikzpicture}[overlay, remember picture]
        \path (current page.north east) ++(-2.1,-0.2) node[below left] {
            This paper has been accepted for publication in the 2024 IEEE International Conference on Robotics and Automation.
        };
    \end{tikzpicture}
}

\newcommand\copyrighttext{%
  \footnotesize \textcopyright 2024 IEEE. Personal use of this material is permitted.
  Permission from IEEE must be obtained for all other uses, in any current or future 
  media, including reprinting/republishing this material for advertising or promotional 
  purposes, creating new collective works, for resale or redistribution to servers or 
  lists, or reuse of any copyrighted component of this work in other works.}
\newcommand\copyrightnotice{%
\begin{tikzpicture}[remember picture,overlay]
\node[anchor=south,yshift=10pt] at (current page.south) {\fbox{\parbox{\dimexpr\textwidth-\fboxsep-\fboxrule\relax}{\copyrighttext}}};
\end{tikzpicture}%
}

\begin{document}
\title{\LARGE \bf Optimal Control for Clutched-Elastic Robots:\\ A Contact-Implicit Approach}
\author{Dennis Ossadnik$^{1,\dagger,*}$, Vasilije Rak\v{c}evi\'c$^{1,\dagger}$, Mehmet C. Yildirim$^{1}$, Edmundo Pozo Fortuni\'{c}$^{1}$, \\  Hugo T. M. Kussaba$^{1}$, Abdalla Swikir$^{1,2}$, and Sami Haddadin$^{1}$
\thanks{$^{1}$The authors are with the Chair of Robotics and Systems Intelligence and the Munich Institute of Robotics and Machine Intelligence (MIRMI), Technical University of Munich, Germany. The authors acknowledge the financial support by the Bavarian State Ministry for Economic Affairs, Regional Development and Energy (StMWi) for the Lighthouse Initiative KI.FABRIK, (Phase 1: Infrastructure as well as the research and development program under, grant no. DIK0249) and European Union's Horizon 2020 research and innovation programme as part of the project Darko under grant no. 101017274, and also under the Marie Skłodowska-Curie grant agreement no. 899987.
$^{*}$Corresponding author: \tt\small{dennis.ossadnik@tum.de}}
\thanks{$^{2}$Abdalla Swikir is also with the Department of Electrical and Electronic Engineering, Omar Al-Mukhtar University (OMU), Albaida, Libya.} 
\thanks{$^{\dagger}$The first two authors contributed equally to this work.}
}

\maketitle
\IEEEacceptance
\copyrightnotice
\thispagestyle{plain}
\pagestyle{plain}

\begin{abstract}
Intrinsically elastic robots surpass their rigid counterparts in a range of different characteristics. By temporarily storing potential energy and subsequently converting it to kinetic energy, elastic robots are capable of highly dynamic motions even with limited motor power. However, the time-dependency of this energy storage and release mechanism remains one of the major challenges in controlling elastic robots. A possible remedy is the introduction of locking elements (i.e. clutches and brakes) in the drive train. This gives rise to a new class of robots, so-called clutched-elastic robots (CER), with which it is possible to precisely control the energy-transfer timing.  A prevalent challenge in the realm of CERs is the automatic discovery of clutch sequences. Due to complexity, many methods still rely on pre-defined modes. In this paper, we introduce a novel contact-implicit scheme designed to optimize both control input and clutch sequence simultaneously. A penalty in the objective function ensures the prevention of unnecessary clutch transitions. We empirically demonstrate the effectiveness of our proposed method on a double pendulum equipped with two of our newly proposed clutch-based Bi-Stiffness Actuators (BSA). %

\end{abstract}

\section{Introduction}
Leveraging the natural dynamics of intrinsically elastic robots (ER) has become a widely embraced paradigm in recent years. Many novel robot designs employ series or parallel elasticity in their drive train. These systems show increased performance in terms of energy efficiency (cf. \cite{Verstraten2016}), peak power (cf. \cite{haddadin2009kick, haddadin2012intrinsically}), and safety (cf. \cite{park2009safe}) in comparison to rigidly actuated robots. Naturally, ERs are associated with some challenges in terms of control. A common issue is the time dependence of energy storage and release \cite{Plooij}. To alleviate this problem, the use of clutches in combination with intrinsic joint elasticity has become increasingly popular. This gives rise to a new class of robots, so-called \textit{Clutched Elastic Robots} (CER), which allow for precise control of the timing of energy storage and release.

For example, in \cite{Leach2014}, a multi-modal actuator is presented, in which dynamic coupling via clutches and brakes is used to produce a wide range of behaviours and controlled energy storage and release. In \cite{haufle2012clutched}, a clutched parallel elastic actuator concept for improved energy efficiency in legged robots is introduced. In our recent work, we also introduced a novel clutched-elastic concept called \textit{Bi-Stiffness Actuation} (BSA, cf. \cite{ossadnik2022BSA}). This concept allows to accurately control the energy transfer timing and can exploit synergistic effects between gravitational and potential energy, as demonstrated by our prototype \cite{Pozo23_RAL}. 
\begin{figure}[t]
    \centering
  	\def\svgwidth{\linewidth}
	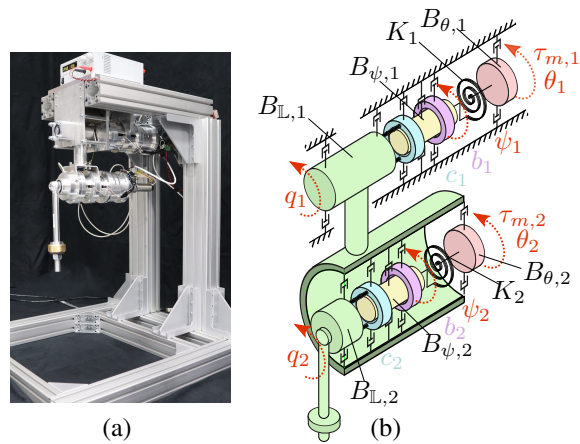     
    \caption{Hardware setup (a): Experimental 2 DoF elastic pendulum actuated by two  Bi-Stiffness Actuators. Internal structure (b): For each joint $j$, the spring inertia can be locked in place by brake $b_j$ (violet) and/or coupled directly to the link by clutch $c_j$ (cyan).}
    \label{fig:BSA_Structure}
    \vspace{-0.5cm}
\end{figure}
\subsection{Optimal Control for Clutched-Elastic Robots (CER)}
Numerous CER prototypes validate their performance through the application of optimal control techniques, often with simplifications, such as providing hard-coded clutching sequences. For example, in \cite{chen2013optimal}, different pre-defined mode sequences are explored, but there is no definitive confirmation that the optimal sequence is included among them.  Similarly, in \cite{plooij2016reducing}, the authors assume optimal switching depending on initial and final positions, though other sequences might yield better results.  In \cite{Krimsky}, a mixed-integer quadratic program for optimization is used. However, it is limited to a linear model and a single actuator scenario. In \cite{ossadnik2022BSA}, trajectory optimization for a clutched elastic double pendulum is considered. Still, only pre-defined mode sequences are used for control. All these examples indicate that the automatic resolution of the mode sequence for the full nonlinear model including multiple degrees of freedom (DoF) still remains an open problem in CERs,  possibly due to the way how clutch contacts are handled.
\subsection{Contact modelling}
CERs are characterized by the interplay of continuous and discrete dynamics due to the rapid change of contact situation as a result of the clutch action and can thus be effectively modelled as hybrid systems. There are two main methods to describe this class of systems \cite{van2007introduction}. More commonly, CERs are modelled using \textit{event-driven} schemes. Here, the continuous dynamics is simulated until an event is detected upon which the new mode is determined and initialized. A common issue of this scheme is the number of possible modes, which increases exponentially with the dimensionality of the system. \textit{Time-stepping methods}, on the other hand, do not rely on event detection but encode the different modes via complementarity constraints \cite{stewart2000implicit}. They scale to high-dimensional systems and reduce the complexity since mode transitions are automatically resolved in the complementarity constraints.
\subsection{Contact-Implicit Optimal Control}
In the legged robotics field, contacts were also usually handled by formulating a multi-stage optimization using event-driven schemes, where contact sequences had to be defined \textit{a priori} by the user (cf. e.g. \cite{mombaur2005open}). In the last decade, however, some very promising optimal control approaches emerged that were able to jointly optimize the control inputs as well as the mode sequence. In \cite{Posa2014}, a time-stepping method is used to formulate a contact-implicit optimal control problem as a nonlinear program (NLP). Instead of predefining the mode sequence, the contact forces are included as decision variables via complementarity constraints. The formulation facilitates dealing with high-dimensional systems with numerous possible mode transitions. This, however, comes at the price of violating some of the constraint qualifications that NLP solvers assume which commonly leads to bad performance (cf. \cite{nurkanovic2020limits}). In practice, a smoothing or relaxation method can be used to alleviate this problem. In \cite{Posa2014}, contact-implicit optimal control could thus be successfully applied to a dynamic locomotion task. Similar formulations have been used for manipulation tasks as well. In \cite{sleiman2019contact}, a contact-implicit scheme is used for dynamic object manipulation with a robotic arm. While these systems share many similarities with CERs, there are some key differences: In case of dynamic locomotion or manipulation, contact is established between the environment and the robot. Thus, switching is \textit{state-based}. In CERs, contacts are used to change the internal state of the robot's actuators. Switching is \textit{user-defined} or \textit{externally forced} (cf. \cite{zhu2015optimal}): That is to say, the user defines a signal $\rho(t): \mathbb{R}^{+} \rightarrow \{0, 1\}$ for each clutch, in which the transitions from 0 to 1 (or vice versa) indicate an instantaneous mode change corresponding to the clutch being engaged or disengaged.
\subsection{Contribution}
To address the characteristic switching behaviour exhibited by CERs, we propose a new contact-implicit scheme for jointly optimizing the control input and the clutch sequence. Our formulation treats the clutch torque as a freely optimizable parameter. This theoretically permits mode changes at each time step. To avoid unnecessary switching, we propose a penalty term in the objective function for shaping the clutch torque signal. To verify our method, we conducted a series of experiments on a clutched-elastic double pendulum that is equipped with two BSAs (see Fig.\ref{fig:BSA_Structure}):
\begin{itemize}
    \item[1.] Optimal control trajectories are generated using both the new method and a pre-defined sequence \cite{ossadnik2022BSA}.
    \item[2.] A closed-loop hybrid LQR controller (adapted from \cite{saccon2014sensitivity} to account for user-defined switching) is applied for trajectory tracking.
\end{itemize}
\section{Modelling} 
\label{sec:modelling}
Modelling CERs necessitates addressing both continuous and discrete dynamics. Initially, we present the continuous dynamics, followed by the introduction of event-driven and subsequently the novel complementarity-based formulation to handle discrete dynamics. %
\subsection{Equation of motion}
\label{sec:preliminaries}
We introduce the vector of link positions $\vect{q}$ as well as the vector of internal states $\vect{\chi}$ and define $\vect{\xi} \coloneqq (\vect{\chi}, \vect{q})$. Similar to \cite{Plooij}, we state the dynamics of a CER as 
\begin{equation}
    \matr{\Pi}(\vect{\xi}) \ddot{\vect{\xi}} + \vect{\eta}(\vect{\xi}, \dot{\vect{\xi}}) + \vect{\tau}_f + \vect{\tau}(\vect{\vect{\xi}, \vect{\tau}_m}) = \vect{\tau}_c, \label{eq:CEA_dynamics} \\ 
\end{equation}
Here, $\matr{\Pi}(\vect{\xi})$ is the generalized mass matrix, $\vect{\eta}(\vect{\xi}, \dot{\vect{\xi}})$ is the nonlinear bias term, $\vect{\tau}_f$ is the friction torque and $\vect{\tau}(\vect{\vect{\xi}, \vect{\tau}_m})$ is the overall internal torque generated by internal springs and motor torque $\vect{\tau}_m$ and $\vect{\tau}_c$ is the torque generated by clutches. %
In this work, we consider a double-pendulum actuated by a specific clutch-based concept, \textit{Bi-Stiffness Actuation} (BSA). Each BSA joint consists of a motor, a spring element in series with the motor, and a link. Each joint is also equipped with mechanical brake $b_i$ and clutch $c_i$, which can either lock the and/or couple it to the link side, which gives rise to the various modes of the actuator. An overview of the system is found in Fig.~\ref{fig:BSA_Structure}. The system parameters are summarized in Tab.~\ref{tab:parameters}.

The internal states are the motor position $\vect{\theta}$ and the spring position $\vect{\psi}$, thus ${\vect{\chi} = (\vect{\theta}, \vect{\psi})}$. The motor dynamics is
\begin{equation}\label{eq:motor_dynamics}
    \matr{B}_{\theta} \ddot{\vect{\theta}} +  \matr{K} (\vect{\theta} - \vect{\psi}) = \vect{\tau}_m.
\end{equation}
As in \cite{Haddadin2012}, we can bring \eqref{eq:motor_dynamics} into singular perturbation form. We assume a control law
$\vect{\tau}_m = \vect{K}_P (\dot{\vect{\theta}}_d - \dot{\vect{\theta}})$, which leads to
\begin{equation}
   \matr{K}_P^{-1} (\matr{B}_{\theta} \ddot{\vect{\theta}} + \matr{K} (\vect{\theta} - \vect{\psi}) ) =  \dot{\vect{\theta}}_d - \dot{\vect{\theta}}.
\end{equation}
\noindent By taking the limit $\matr{K}_P^{-1} \rightarrow \vect{0}$, we obtain $\dot{\vect{\theta}} = \dot{\vect{\theta}}_d$. That is to say, we can assume the motor velocity as our new control input $\vect{u} \coloneqq \dot{\vect{\theta}}_d$. Thus, the motor position can be removed and the notation is simplified, i.e. $\vect{\chi} = \vect{\psi}$.
Now, as in \cite{ossadnik2022BSA}, the continuous dynamics for BSA can be formulated as
\begin{equation}
    \underbrace{
    \begin{bmatrix} 
	        \matr{B}_{\psi} & \vect{0} \\
	        \vect{0}  & \matr{M}(\vect{q})
	\end{bmatrix}}_{= \matr{\Pi}(\vect{\xi})} \ddot{\vect{\xi}} + \underbrace{
    \begin{bmatrix}
        \vect{0} \\
	    \vect{h}(\vect{q}, \dot{\vect{q}})
	\end{bmatrix}}_{= \vect{\eta}(\vect{\xi}, \dot{\vect{\xi}})} + \underbrace{
	\begin{bmatrix}
        \matr{K} (\vect{\theta} - \vect{\psi}) \\
        \vect{0}
    \end{bmatrix}}_{= \vect{\tau}(\vect{\xi})} + \vect{\tau}_f = \vect{\tau}_c, 
    \label{eq:robot}
\end{equation}
Here, $\matr{B}_{\theta}$, $\matr{B}_{\psi}$, $\matr{M}(\vect{q})$, $\matr{K}$, and $\vect{h}(\vect{q}, \dot{\vect{q}})$ denote the motor, spring and link-side inertias, the stiffness matrix, and the link-side nonlinear bias term,  respectively.
In each joint, there are two clutches: one can lock or unlock the spring element, the other couples or decouples the link from the spring. If a clutch is engaged, the relative speed between the two frames it connects becomes zero. 
We define the vector of relative speeds 
\begin{equation}
    \vect{\varphi} \coloneqq \left( \dot{\psi}_1, \dot{\psi}_1 - \dot{q}_1, \dot{\psi}_2, \dot{\psi}_2 - \dot{q}_2\right)
\end{equation}
and the index set
\begin{gather}
    \mathcal{I} \coloneqq \{i \in \{1,2,3,4\} \,|\, \varphi_i = 0\},
\end{gather}
where $ \varphi_i$ is the $i$-th element of $ \vect{\varphi}$. Next, we define the vector 
\begin{equation}
    \vect{\vartheta} \coloneqq \left( \varphi_i \right)_{i \in \mathcal{I}} \label{eq:relspeeds_concat}
\end{equation}
which contains the relative speeds of all engaged clutches.
In case of BSA, this leads to four possible modes per joint, which are summarized in Tab.~\ref{tab:modes}. Here, the vector $\vect{\vartheta}_j$ contains the relative speeds of all active clutches in joint $j$. The full vector is obtained as $\vect{\vartheta} = \left(\vect{\vartheta}_1, \vect{\vartheta}_2\right)$. We will introduce two modelling techniques to compute $\vect{\tau}_c$. First, we recapitulate the event-driven scheme presented in our previous work. Next, we introduce a time-stepping formulation that will be used to formulate the optimal control problem. 

\subsection{Event-driven formulation} 
We express \eqref{eq:relspeeds_concat} as $\matr{C}_p \dot{\vect{\xi}} = \vect{\vartheta}$, 
where $\matr{C}_p \coloneqq {\partial \vect{\vartheta}}/{\partial \dot{\vect{\xi}}}$. Setting $\vect{\tau}_c = \matr{C}_p^{\mathsf{T}} \vect{\lambda}$, differentiating and rearranging leads to a direct expression for the constraint torque (cf. \cite{ossadnik2022BSA})
    \begin{equation} \label{eq:constr_torque}
		 \vect{\lambda} = (\matr{C}_p \matr{\Pi}^{-1} \matr{C}_p^{\mathsf{T}})^{-1}\matr{C}_p \matr{\Pi}^{-1} ( \vect{\tau} + \vect{\eta}).
\end{equation} 
When a change in the contact situation occurs, the new constraint is enforced by an instantaneous impulse
\begin{equation}
\begin{aligned}
	\dotxik{+} &= \dotxik{-} +  \matr{\Pi}^{-1} \matr{C}_p^{\mathsf{T}} \vect{\Lambda}, \\ 
	\vect{\Lambda} &= -(\matr{C}_p \matr{\Pi}^{-1} \matr{C}_p^{\mathsf{T}})^{-1} \matr{C}_p \dotxik{-}.    \label{eq:impact_law}
\end{aligned}
\end{equation}
Here, $\dotxik{-}$ (resp. $\dotxik{+}$) denotes the velocity right before (resp. after) the impact, and $\vect{\Lambda}_p$ denotes the contact impulse. Introducing the state vector ${\vect{x} \coloneqq (\vect{\theta}, \vect{\xi}, \dot{\vect{\xi}})}$, we obtain the continuous dynamics
\begin{equation}
	\dot{\vect{x}} = \vect{f}_p(\vect{x}, \vect{u}) \coloneqq \begin{bmatrix}
	\vect{u} \\
	\dot{\vect{\xi}} \\
	 \matr{\Pi}^{-1} (\matr{C}_p^{\mathsf{T}} \vect{\lambda} - \vect{\eta} - \vect{\tau})
	\end{bmatrix}. \label{eq:dyn}
\end{equation}
At a switching instance, the reset map $\vect{g}_p$ has to be evaluated:
\begin{equation}
\vect{x}^{+} = \vect{g}_p(\vect{x}^{-}) \coloneqq \begin{bmatrix}
\vect{\theta}^{-} \\
\vect{\xi}^{-} \\
\dotxik{-} + \matr{\Pi}^{-1} \matr{C}_p^{\mathsf{T}} \vect{\Lambda} 
\end{bmatrix}. \label{eq:jump}
\end{equation}
In an optimal control setting, an event-driven formulation usually requires a predefined sequence of dynamic modes. Using a contact-implicit formulation, the modes can be encoded in the constraints. Thus, the sequence can be optimized alongside the control input. \begin{table}[t]
\vspace{0.2cm}
\caption{Actuator modes for the $j$-th joint. DEC (Decoupled), SEA, STG (Storage), BRK (Braked Storage).  If $c_j = 1$, $b_j = 1$ the clutch and brake are engaged, else if  $c_j = 0$, $b_j = 0$, disengaged.} \label{tab:modes}
\vspace{-0.3cm}
\begin{minipage}{\linewidth}
\begin{minipage}{0.65\linewidth}
\centering

    \begin{tabular}[c]{|c|c|c|c|c|} 
	\hline	\textbf{Mode} & $b_j$ & $c_j$ &  $\vect{\vartheta}_j$\\
	\hline \rule{0pt}{4ex}    DEC &0& 0& $- $        ~\\[0.25cm]
	\hline	\rule{0pt}{4ex}    SEA &0& 1& $\dot{\psi}_j - \dot{q}_j$ ~\\[0.25cm]
 	\hline \rule{0pt}{4ex}    STG  &1& 0& $ \dot{\psi}_j $        ~\\[0.25cm]
 	\hline \rule{0pt}{4ex}    BRK  &1& 1& $ \begin{bmatrix}
 	      \dot{\psi}_j  \\
	\dot{\psi}_j - \dot{q}_j
 	\end{bmatrix}
$        ~\\[0.25cm]
 \hline
\end{tabular}
\end{minipage} \begin{minipage}{0.34\linewidth}
 \def\svgwidth{1.\linewidth}
 \vspace*{0.05cm}
	\hspace*{-0.3cm} \begingroup%
  \makeatletter%
  \providecommand\color[2][]{%
    \errmessage{(Inkscape) Color is used for the text in Inkscape, but the package 'color.sty' is not loaded}%
    \renewcommand\color[2][]{}%
  }%
  \providecommand\transparent[1]{%
    \errmessage{(Inkscape) Transparency is used (non-zero) for the text in Inkscape, but the package 'transparent.sty' is not loaded}%
    \renewcommand\transparent[1]{}%
  }%
  \providecommand\rotatebox[2]{#2}%
  \newcommand*\fsize{\dimexpr\f@size pt\relax}%
  \newcommand*\lineheight[1]{\fontsize{\fsize}{#1\fsize}\selectfont}%
  \ifx\svgwidth\undefined%
    \setlength{\unitlength}{146.00951187bp}%
    \ifx\svgscale\undefined%
      \relax%
    \else%
      \setlength{\unitlength}{\unitlength * \real{\svgscale}}%
    \fi%
  \else%
    \setlength{\unitlength}{\svgwidth}%
  \fi%
  \global\let\svgwidth\undefined%
  \global\let\svgscale\undefined%
  \makeatother%
  \begin{picture}(1,1.66439464)%
    \lineheight{1}%
    \setlength\tabcolsep{0pt}%
    \put(0,0){\includegraphics[width=\unitlength,page=1]{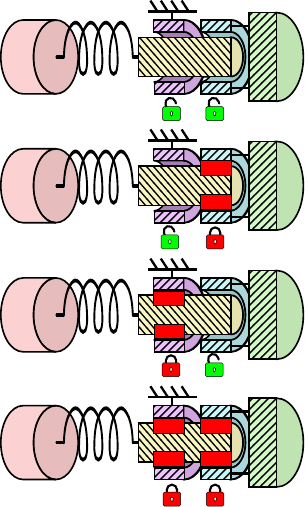}}%
  \end{picture}%
\endgroup%
 
\end{minipage}

\end{minipage}
\vspace{-0.4cm}
\end{table}

\subsection{Time-stepping formulation}
Applying the backward Euler method (cf. \cite{Posa2014}) with time-step length $\delta$ to  \eqref{eq:CEA_dynamics} and setting $\vect{\tau}_c = \sum_{i=1}^{4}\matr{\Gamma}_i^{\mathsf{T}} {\zeta}_i^{k+1}$, where \mbox{$\matr{\Gamma}_i \coloneqq {\partial \varphi_i}/{\partial \dot{\vect{\xi}}}$} and ${\zeta}^{k+1}_i$ is the associated friction torque, the following discrete dynamics are obtained: 
\begin{equation}
\begin{gathered}
\vect{\xi}^{k+1} = {\vect{\xi}}^{k} + \delta \dotxik{k+1}\\
\matr{\Pi}^{k+1} (\dotxik{k+1} - \dotxik{k}) + \delta (\vect{\eta}^{k+1} + \vect{\tau}^{k+1}) = \sum_{i=1}^{4}\nolimits\matr{\Gamma}_i^{\mathsf{T}} {\zeta}_i^{k+1}, \label{eq:disc_dyn2}
\end{gathered}
\end{equation}
where $\matr{\Pi}^{k+1} \coloneqq \matr{\Pi}(\vect{\xi}^{k+1})$, $\vect{\eta}^{k+1} \coloneqq \vect{\eta}(\vect{\xi}^{k+1}, \dotxik{k+1})$, and $\vect{\tau}^{k+1} \coloneqq \vect{\tau}(\vect{\xi}^{k+1})$.
Rearranging \eqref{eq:disc_dyn2} results in
\begin{equation}    
    \dotxik{k+1} = 
    (\matr{\Pi}^{k+1}){\vphantom{(\matr{\Pi})}}^{\!{}^{-1}}\!\!
    \left[ \sum_{i=1}^{4}\nolimits\matr{\Gamma}_i^{\mathsf{T}}{\zeta}^{k+1}_{i}\!-\! \delta (\vect{\eta}^{k+1} + \vect{\tau}^{k+1}) \right]\!+\!\dotxik{k}.
\end{equation}
Introducing ${\vect{x}^k \coloneqq (\vect{\theta}^k, \vect{\xi}^{k}, \dotxik{k})}$,  $\vect{u}^k \coloneqq \dotthetak{k}$ and $\vect{\zeta}^k \coloneqq (\zeta^1_k, \dots, \zeta^4_k)$, we can formulate the discrete-time dynamics more compactly as
\begin{equation} \label{eq:disc_dynamics}
   \vect{x}^{k+1} = \vect{f}(\vect{x}^k, \vect{u}^k, \vect{\zeta}^k)\coloneqq \begin{bmatrix}
      \vect{u}^k \\
      {\vect{\xi}}^{k} + \delta \dotxik{k+1}   \\
      \dotxik{k+1} 
  \end{bmatrix}.
\end{equation}
To compute the clutch torque, we employ a complementarity formulation. The friction torque is split into positive and negative components
\begin{equation} \label{eq:lambda_def}
    {\zeta}_i^k = {\pi}_i^k - {\nu}_i^k.
\end{equation}
Similar to \cite{Posa2014}, using an additional slack variable ${\gamma}_i^k \geq 0$, the following conditions hold:\footnote{The symbol $\perp$ denotes a complementarity relation, that is, $a \perp b$ if and only if $a b = 0, a,b \geq 0$. In practice, the complementarity relation is relaxed to $a b \leq \epsilon$, where $\epsilon > 0$ is a relaxation parameter. The problem is then solved multiple times with decreasing values for $\epsilon$ using the solution of the last iteration as a warm start.}
\begin{align}
    { (\gamma}_i^k + {\varphi_i^k}) &\perp  {\pi}_i^k, \label{eq:comp1}\\
    { (\gamma}_i^k - {\varphi_i^k}) &\perp  {\nu}_i^k, \label{eq:comp2}
\end{align}
 In contrast to the formulation presented in \cite{Posa2014}, where contact is established solely when the distance between a contact point on the robot and the environment reaches zero, there is no distance function in our case. Instead, contact can be initiated in every time step when the optimizer selects a non-zero $\vect{\zeta}^k$. \\

\section{Optimal Control} \label{sec:OC}
In this section, we present a contact-implicit optimization scheme based on the previously formulated time-stepping approach and the objective function design to generate open-loop trajectories. 
\subsection{Trajectory optimization}
For formulating the optimization problem, the clutch torque $\vect{\zeta}^k$ is exposed as an active decision variable. Hereby, the optimizer can reason about the optimal contact sequence, i.e., to determine when to engage and disengage the clutches in the drive train. The optimization problem can be stated as
\begin{equation}
    \label{eq:OCP_traj_opt}
    \begin{gathered}
        \underset{\vect{x}^k, \vect{u}^k, \vect{\zeta}^k}{\text{minimize}}  \quad \mathcal{J}(\vect{x}^k, \vect{u}^k,\vect{\zeta}^k) \\
        \text{s.t.} \begin{Bmatrix*}[l]
        \text{Eq. \eqref{eq:disc_dynamics}} & \text{Dynamics} \\
        \text{Eqs. \eqref{eq:lambda_def}--\eqref{eq:comp2}} & \text{Contact handling} \\
        \vect{x}^k \in \mathcal{X} & \text{State constraints} \\
        \vect{u}^k \in \mathcal{U} & \text{Control constraints}
        \end{Bmatrix*}.
    \end{gathered}
\end{equation}
Next, we will explain the formulation of the objective function, which has to be designed in a specific way to take into account the freely optimizable clutch torque. 
\subsection{Objective function}
When designing the objective function, we have several goals in mind. The main goal is speed maximization, specifically focusing on the end-links velocity $\vect{v}_{EE} \coloneqq \vect{J}(\vect{q}^n) \dot{\vect{q}}^n $ at the final time $T = n\delta$. %
The term in the objective function corresponding to this goal can be expressed as
\begin{equation}
    \mathcal{J}_1(\vect{x}^k) \coloneqq -\lVert \vect{v}_{EE} \rVert ^2.
\end{equation}
The optimal control formulation so far would allow switching modes in every time step. We seek a signal that encodes the number of mode changes such that we can formulate a penalty term for it. For this reason, we first define an indicator signal that corresponds to the clutch being engaged and disengaged. At each step $k$, we define the indicator signal
\begin{equation}
    \sigma_{i}^k \coloneqq e^{-\alpha {\zeta_{i}^k\zeta_{i}^k}} - \frac{1}{2}  \approx \begin{cases} \frac{1}{2}, &\text{if } \zeta_{i}^k = 0, \\
    -\frac{1}{2}, &\text{otherwise,}
    \end{cases} 
\end{equation}
where $\alpha >0$ is a smoothing parameter. The signal serves as a classifier, which is positive, if clutch $j$ is disengaged at the current step $k$, and negative, if it is engaged. The number of zero crossings in this signal corresponds to the number of switching instances, making it a suitable penalty term in the formulation of the cost function. This count can be approximated as:\footnote{Here, a smooth approximation for the sign function is used: \mbox{$\tanh( \beta \bullet) \approx \mathrm{sign}(\bullet)$)}, where $\beta >0$ is a smoothing parameter.}
\begin{equation}
    \mathcal{J}_{2}(\vect{\zeta}^k) \coloneqq  \sum_{i=1}^4\nolimits \sum_{k=2}^n\nolimits \frac{1}{2} (1 + \tanh(\beta  (- \sigma_{i}^{k-1}   \sigma_{i}^k ))).
\end{equation}
Additionally, we add a regularization term $\mathcal{J}_{3}(\vect{u}^k) \coloneqq \sum_{k=1}^n \lVert \vect{u}^{k} \rVert ^2$.
The objective function can be defined as the weighted sum of our main objective (speed maximization), the switching penalty and the regularization term, i.e.
\begin{equation}
    \mathcal{J}(\vect{x}^k,\vect{u}^k, \vect{\zeta}^k) \coloneqq w_1 \mathcal{J}_1(\vect{x}^k) + w_2 \mathcal{J}_{2}(\vect{\zeta}^k) + w_3 \mathcal{J}_{3}(\vect{u}^k),
\end{equation}
where $w_1, w_2, w_3 >0$ are appropriate weighting factors.

\section{Trajectory tracking} \label{sec:tracking}

To respond to system disturbances and accommodate even minor fluctuations in system parameters, we implemented a closed-loop feedback controller to track the trajectory $(\dot{\vect{\theta}}_\textrm{ref}(t), \vect{\xi}_\textrm{ref}(t))$ resulting from a numerical interpolation of the solution of \eqref{eq:OCP_traj_opt}.

As shown in \cite{saccon2014sensitivity} and references therein, small variations in initial conditions lead to variations on the jumping solution of a hybrid system, making the tracking problem of the optimal clutch sequence more challenging than the tracking problem in purely continuous-time systems.  
Nevertheless, the classic LQR controller for continuous-time systems can be generalized to a hybrid setting \cite{saccon2014sensitivity}. 
More precisely, let $\matr{Q}$, $\matr{R}$ and $\matr{P}_T$ be positive-definite matrices with appropriate dimensions, and consider the minimization problem of the functional
\begin{equation}
\frac{1}{2} \int_0^T \vect{x}^\mathsf{T}(s) \matr{Q} \vect{x}(s)+\vect{u}^\mathsf{T}(s) \matr{R} \vect{u}(s) \, ds+\frac{1}{2} \vect{x}^\mathsf{T}(T) \matr{P}_T \vect{x}(T)    
\end{equation}
over the horizon $[0,T]$, with $N$ switches happening during the instants $\tilde{t}^i$, $i=1,\ldots, N-1$. For the sake of notation convenience, we define $\tilde{t}^0 := 0$ and $\tilde{t}^N := T$.

The hybrid-feedback law to track $(\dot{\vect{\theta}}_\textrm{ref}(t), \vect{\xi}_\textrm{ref}(t))$ is given by \cite{saccon2014sensitivity} as
\begin{equation}\label{eq:hybrid_LQR}
    \vect{u}(\vect{\xi}, t) := \dot{\vect{\theta}}_\textrm{ref}(t) - \matr{R}^{-1} \BpT
    \matr{P}(t)(\vect{\xi} - \vect{\xi}_\textrm{ref}(t)),
\end{equation}
where $\matr{P}(t)$ is defined piecewise by solving (backward in time) each of the following Riccati differential equations
for $ t \in (\tilde{t}^k, \tilde{t}^{k+1}]$, $k = N-1, \ldots, 0$:
\begin{gather*}
    -\dot{\matr{P}}(t)\!=\!\ApT \matr{P}(t)+\matr{P}(t)\Ap-\matr{P}(t)\matr{S}(t)\matr{P}(t)+\matr{Q},
\end{gather*}
with $\matr{S}(t):= (\iBp) \matr{R}^{-1} (\iBp)^\mathsf{T}$, and associated terminal boundary conditions given by
\begin{align*}
    \matr{P}(T)   &= \matr{P}_{T}, \\
    \matr{P}(\tilde{t}^k) &= 
    (I + \matr{H}_{q,p})^{\mathsf{T}} \matr{P}(\tilde{t}^{k+1}) (\matr{I} + \matr{H}_{q,p}), \, k\!\!=\!\!1,\ldots,N\!\!-\!\!2,
\end{align*}
with $\matr{H}_{q,p}(\tilde{t})$ being the sensitivity of the $p$-th mode of the system over a jump \cite{saccon2014sensitivity} to the $q$-th mode at the instant $\tilde{t}$ and given by
\begin{gather}\label{eq:generic_jump_sensitivity}
\matr{H}_{q,p}(\tilde{t}) = \matr{G}_{q,p}(\tilde{t})
  \frac{\partial \vect{g}_p}{\partial \vect{x}}(\vect{x}^-) 
  + %
   \frac{\partial \matr{\Delta}^{-}}{\partial \vect{x}}(\vect{x}^-),
\end{gather}
where
\begin{align*}
\matr{G}_{q,p}(\tilde{t}) &:=
\frac{\vect{f}_q\left(\vect{x}^{+}, \vect{u}\right)-\vect{f}_p\left(\vect{x}^{-}, \vect{u}\right)-\dot{\vect{\Delta}}^{-}\left(\vect{x}^{-}, \vect{u}\right)}{\frac{\partial \vect{g}_p}{\partial \vect{x}}\left(\vect{x}^{-}\right) \vect{f}_p\left(\vect{x}^{-}, \vect{u}\right)+\frac{\partial \vect{g}_p}{\partial t}\left(\vect{x}^{-}\right)}, \\
\matr{\Delta}^{-}(\vect{x}) &:= \vect{g}_p(\vect{x}) - \vect{x}, \\
\dot{\matr{\Delta}}^{-}(\vect{x}^-,\vect{u}) &:= \frac{\partial \matr{\Delta}^{-}}{\partial \vect{x}}(\vect{x}^-)\vect{f}_p(\vect{x}^-,\vect{u})  - \frac{\partial \matr{\Delta}^{-}}{\partial \vect{x}}(\vect{x}^-).
\end{align*}
For systems without state-dependent switches such as the BSA, the first term of \eqref{eq:generic_jump_sensitivity} is null and 
$$
\matr{H}_{q,p}(\tilde{t}) = \matr{H}_p(\tilde{t}) := \frac{\partial}{\partial \vect{x}}(\vect{g}_p(\vect{x}) - \vect{x}).
$$

\section{Experiments}\label{sec:results}
\begin{figure*}[ht]
    \vspace{0.2cm}
    \centering
  	\def\svgwidth{\linewidth}
   	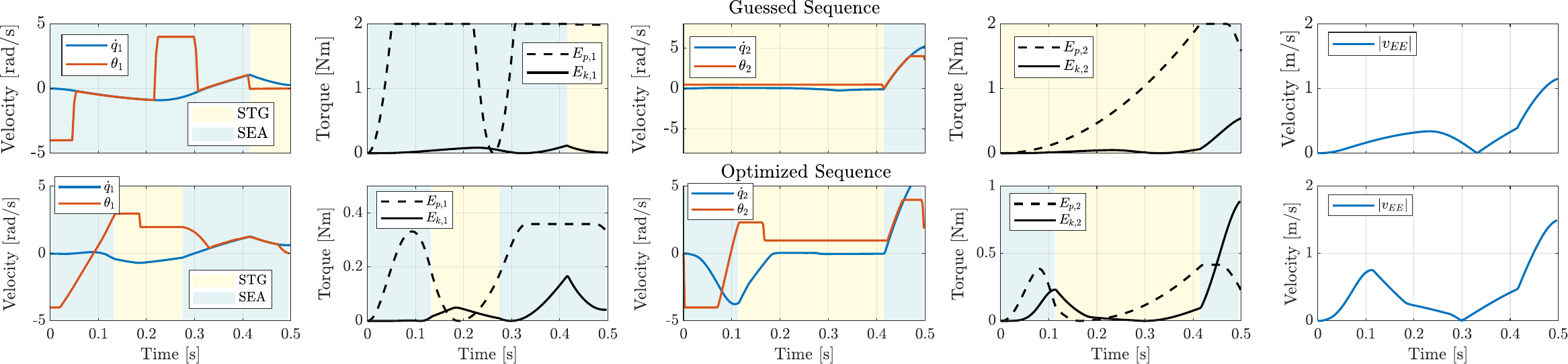 
 \vspace{-0.5cm}
    \caption{Optimization results - \textit{Speed maximization at fixed final time}. The upper and lower rows show the “guessed” and optimized sequence, respectively. The background colour indicates the operating mode (yellow for STG and turquoise for SEA)}
    \label{fig:opt_old_new}
     \vspace{-0.3cm}
\end{figure*}
\begin{figure*}[ht]
    \vspace{0.2cm}
    \centering
  	\def\svgwidth{\linewidth}
	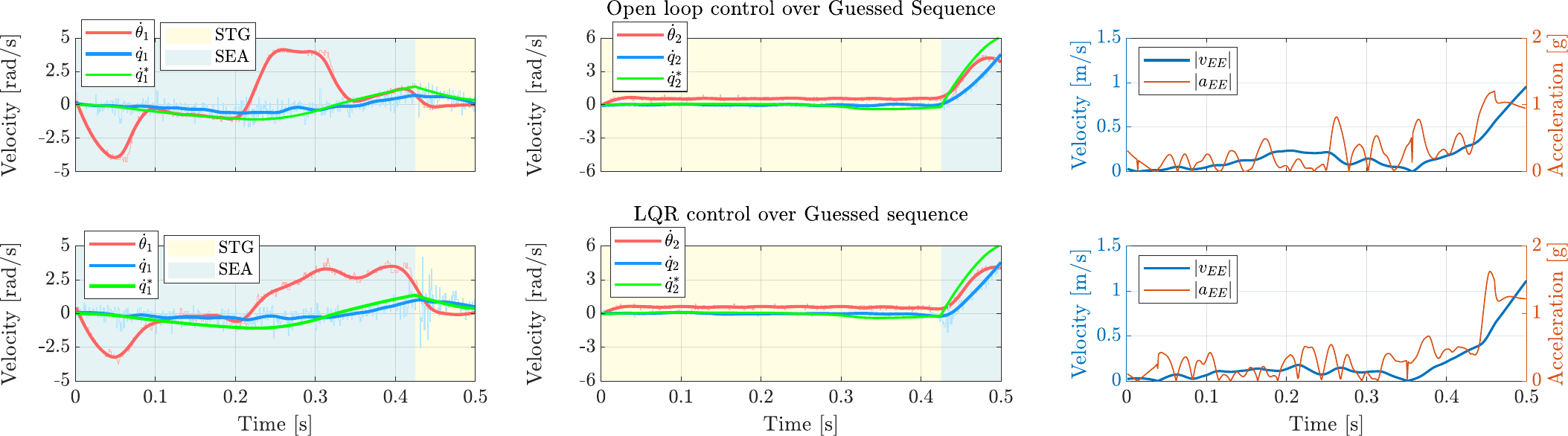 
 \vspace{-0.5cm}
    \caption{Experimental results - \textit{Speed maximization at fixed final time -- Guessed Sequence}. The upper and lower rows show the results with and without LQR controller. The background colour indicates the operating mode (yellow for STG and turquoise for SEA).  
    In the left and middle plots, you can observe the changes in motor angle $\dot{\theta}_{1,2}$ (red) and link angle $\dot{q}_{1,2}$ (blue) as raw and filtered data and the reference $\dot{q}_{1,2}^*$. The data smoothing procedure involved the application of robust quadratic regression, performed utilizing Matlab's smoothdata function. The plots on the right show the magnitude of the end effectors velocity $\vert v_{EE} \vert$ (blue) and acceleration $\vert a_{EE} \vert$ (red).}
    \label{fig:exp_guess}
    \vspace{-0.3cm}
\end{figure*}
\begin{figure*}[ht]
    \vspace{0.2cm}
    \centering
  	\def\svgwidth{\linewidth}
	\input{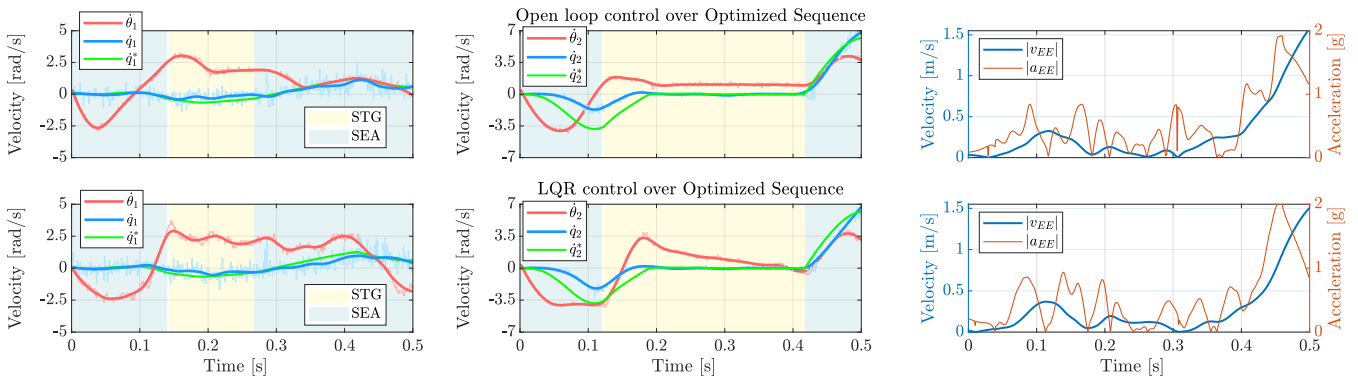} 
    \vspace{-0.5cm}
    \caption{Experimental results - \textit{Speed maximization at fixed final time -- Optimized Sequence}. The plots are arranged as in Fig.~\ref{fig:exp_guess}.}
    \label{fig:exp_opt}
    \vspace{-0.5cm}
\end{figure*}

CER display a wide range of dynamic behaviours, some of which may be unexpected. Sequences based on intuition or heuristics might be suboptimal.  Some resulting sequences might not have been taken into account before. Therefore, in this section, we conduct a series of experiments, in which we compare the results generated by the new method with the results obtained by the “guessed” sequence from our previous work \cite{ossadnik2022BSA}. As an experimental platform, we use a double-pendulum that is actuated by two BSAs. In the following, we denote the first and second joint as J$_{1}$ and  J$_{2}$ respectively.
\subsection{Setup implementation details}

  \begin{figure}[ht]
    \vspace{0.2cm}
    \centering
  	\def\svgwidth{\linewidth}
	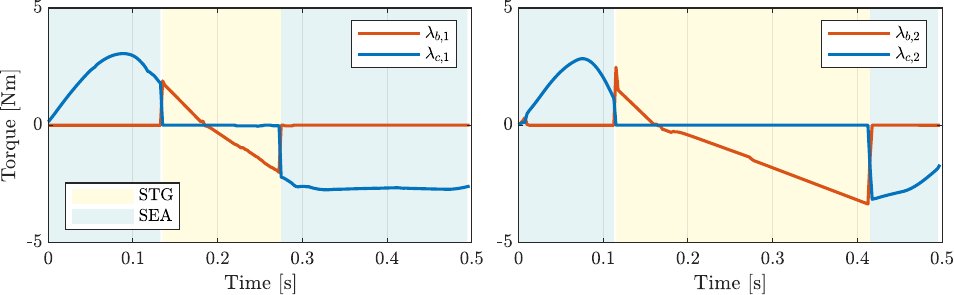 
 \vspace{-0.4cm}
    \caption{Clutch torques generated by contact-implicit optimal control. The torques correspond to the STG and SEA modes, indicated by the background colour.}
    \label{fig:explanation}
    \vspace{-0.7cm}
\end{figure}

Our double pendulum system consists of two configurable modular subsystems that correspond to the prototype presented in \cite{Pozo23_RAL}. J$_2$ utilizes a revised version with optimized weight. The torque sensors in both actuators have been upgraded to the $\pm$ 10 Nm range.
A control PC (x64 running an Ubuntu 22.04) hosts the high-level real-time control routine using a Matlab / Simulink environment (MathWorks, MA, USA) and the EtherCAT master controller using Etherlab (Ingenieurgemeinschaft IgH, Germany) at 1 kHz. The optimal control problem is formulated in CasADi \cite{Andersson2019} and solved using Ipopt \cite{wachter2006implementation} within the Matlab environment (outside the real-time system) using our system model (Sec.\ref{sec:modelling}) and system parameters defined in Tab.\ref{tab:parameters}. The resulting trajectories are stored and used as look-up tables (LUTs). %
Our high-level control routine is comprised of an online simulation block using our system model and parameters that allow us to compare it with our real-world acquired data, and a main controller that handles the operational modes, reading of feedback signals, and selection of desired set points from the trajectory LUTs. %

\subsection{Speed maximization at fixed final time}

\begin{table}[t]
\centering
\vspace*{0.3cm}
        \caption{System parameters of the BSA double pendulum prototype.} 
    \label{tab:parameters}
\begin{tabular}{|l|c|c|c|c|c|}
    \hline
\multirow{2}{*}{Parameter}  & \multirow{2}{*}{Symbol} & \multirow{2}{*}{Unit} & \multicolumn{2}{c|}{Joint values}	\\ \cline{4-5}
                            &                         &                       & $j = 1$     & $j=2$ \\ \hline 
Motor Inertia		        & $B_{\theta,i}$          & $\mathrm{kgm^2}$	  &\multicolumn{2}{c|}{2.38e-5}      \\ \hline 
Spring  inertia	            & $B_{\psi,i}$            & $\mathrm{kgm^2}$	  & 6.20e-4 & 6.15e-4   \\ \hline 
Link Inertia                & $B_{\mathbb{L},i}$      & $\mathrm{kgm^2}$	  & 1.12e-1     & 4.4e-3 \\ \hline 
Link Mass                   & $m_{\mathbb{L},i}$      & $\mathrm{kg}$	      & 9.92e0         & 9.5e-1\\ \hline
Link CoM                    & $r_{\mathbb{L},i}$      & $\mathrm{m}$	      & 9.87e-2         &  1.74e-1\\ \hline
Link Coulomb friction       & $\tau_{C, q,i}$         & $\mathrm{Nm}$	      &  2.00e-1      & 6.00e-1\\ \hline
Link viscous damping        & $d_{q,i}$               & $\mathrm{kgm^2/s}$    & 1.00e-1      & 8.00e-2 \\ \hline
\begin{tabular}[l]{@{}l@{}}Spring Coulomb\\ friction\end{tabular}      & $\tau_{C, \psi,i}$      & $\mathrm{Nm}$	      &\multicolumn{2}{c|}{2.00e-1 } \\ \hline
\begin{tabular}[l]{@{}l@{}}Spring viscous\\ damping\end{tabular}   & $d_{\psi,i}$            & $\mathrm{kgm^2/s}$    & \multicolumn{2}{c|}{1.00e-1 } \\ \hline
Motor Torque                & ${\tau}_m$              & $\mathrm{Nm}$	      & \multicolumn{2}{c|}{$\pm 10$} \\ \hline
Joint Angle Range           & $\theta_{\mathrm{max}}$ & $\mathrm{rad}$	      & \multicolumn{2}{c|}{$\pm 1.2$}         \\ \hline
Max. Spring Deflection      & $\phi_{\mathrm{max}}$   & $\mathrm{rad}$	      & \multicolumn{2}{c|}{$\pm 0.3$}   \\ \hline
\begin{tabular}[l]{@{}l@{}}Max. allowed \\ Spring torque \end{tabular}          & $\tau_{s,\mathrm{max}}$ & $\mathrm{Nm}$	      & $\pm 3.75$   & $\pm 4.35$   \\ \hline

Spring Stiffness            & $K_i$                   &$\mathrm{Nm/rad}$      & $12.5$        & $14.5$ \\ \hline

    \end{tabular}

\end{table}

For evaluation, we compare our new method, which is jointly optimizing both control input and the clutch sequence, to the method used in our previous work \cite{ossadnik2022BSA}, in which a pre-defined clutch sequence is used. Similar to that work, we set the final time to $T=0.5$s and optimize for maximum speed. The correspondence of the clutch torques generated by the contact-implicit scheme with the modes can be seen in Fig.~\ref{fig:explanation}. Fig.~\ref{fig:opt_old_new} shows the trajectories in case of the “guessed” contact sequence and an optimized sequence obtained by the contact-implicit scheme. \\
\textbf{Optimized Sequence.} In case of the optimized sequence, we can observe that both actuators start in SEA mode and simultaneously move their respective links. Then, at $t = 0.11$s and $t = 0.14$s, first  J$_{2}$ and then J$_{1}$ switch to STG mode and preload the springs in the opposite direction. Both links are now fully decoupled and move passively. 
At $t = 0.27$s, J$_{1}$ switches back to SEA mode and moves the first link forward. Due to spring deflection limits, it is not possible to fully convert the stored potential energy to kinetic energy. With some delay, at $t = 0.41$s, J$_{2}$ switches to SEA mode as well. Here, we can observe an almost complete conversion potential to kinetic energy leading to a final velocity of $\| \vect{v}_{EE} \| = 1.5$ m/s. \\
\textbf{Guessed Sequence.} In the guessed sequence, J$_{1}$ starts in SEA mode, while J$_{2}$ is in STG mode preloading the spring. J$_{1}$ uses a resonant swing-up motion and passively excites the second link with it. At $t = 0.41$s, J$_{1}$ switches to DEC, and Joint 2 switches to SEA. J$_{2}$ now swings the link forward, leading to a final velocity of $\| \vect{v}_{EE} \| = 1.1$ m/s.\\
For both sequences, each joint reaches its maximum velocity in perfect succession. The motion thus closely resembles the so-called \textit{proximo-distal sequence} known from biomechanics (cf. \cite{heger2021biomechanische}). However, in the optimized switching sequence, a countermovement in the opposite direction is performed during the first part of the motion, leading to an improvement of 30\% in terms of the reached end-velocity.

\subsection{Experimental results}
The optimal trajectories from the previous section have been tested on a 2DoF double pendulum system integrating two BSA joints. The results can be seen in Fig.~\ref{fig:exp_guess} and Fig.~\ref{fig:exp_opt}. The real system exhibits parameter discrepancies and unmodeled nonlinear effects. Therefore, to be able to follow the trajectory generated by the optimizer, the LQR controller described in Sec.~\ref{sec:tracking} is deployed. The open-loop and closed-loop behaviour is shown for each sequence. Since the ``guessed" sequence already uses most of the available motor speed (which is satured at 4.5 rad/s), higher entries in $\matr{R}$ had to be used to make the control input stay within the limits. The maximum tracking error is 5.2412. On the other hand the optimized sequence used less effort, and thus, the gains in $\matr{R}$ could be chosen less conservatively. Consequently, the tracking performance is significantly improved: the maximum tracking error is decreased to 2.515.
\section{Conclusion}
\label{sec:conclusion}
In this paper, we presented a novel contact-implicit optimal control approach for CER enabling simultaneous optimization of control inputs and clutch sequence. Our method addresses the externally forced switching characteristic of CER and allows to discover new and (possibly) unintuitive clutch sequences, leading to improved performance. The experimental results demonstrate the effectiveness of our approach in comparison to a fixed mode sequence.

\balance

\setlength{\baselineskip}{0pt}


\begin{thebibliography}{10}

\bibitem{Verstraten2016}
T.~Verstraten, P.~Beckerle, R.~Furnémont, G.~Mathijssen, B.~Vanderborght, and D.~Lefeber, ``Series and parallel elastic actuation: Impact of natural dynamics on power and energy consumption,'' {\em Mechanism and Machine Theory}, vol.~102, pp.~232--246, 2016.

\bibitem{haddadin2009kick}
S.~Haddadin, T.~Laue, U.~Frese, S.~Wolf, A.~Albu-Sch{\"a}ffer, and G.~Hirzinger, ``Kick it with elasticity: Safety and performance in human--robot soccer,'' {\em Robot. and Autonomous Syst.}, vol.~57, no.~8, pp.~761--775, 2009.

\bibitem{haddadin2012intrinsically}
S.~Haddadin, F.~Huber, K.~Krieger, R.~Weitschat, A.~Albu-Sch{\"a}ffer, S.~Wolf, W.~Friedl, M.~Grebenstein, F.~Petit, J.~Reinecke, {\em et~al.}, ``Intrinsically elastic robots: The key to human like performance,'' in {\em Proc. 2012 IEEE/RSJ Int. Conf. on Intell. Robots and Syst.}, pp.~4270--4271, 2012.

\bibitem{park2009safe}
J.-J. Park, H.-S. Kim, and J.-B. Song, ``Safe robot arm with safe joint mechanism using nonlinear spring system for collision safety,'' in {\em Proc. 2009 IEEE Int. Conf. on Robot. and Autom.}, pp.~3371--3376, 2009.

\bibitem{Plooij}
M.~Plooij, W.~Wolfslag, and M.~Wisse, ``Clutched {E}lastic {A}ctuators,'' {\em IEEE Trans. Mechatron.}, vol.~22, no.~2, pp.~739--750, 2017.

\bibitem{Leach2014}
D.~Leach, F.~Günther, N.~Maheshwari, and F.~Iida, ``Linear multimodal actuation through discrete coupling,'' {\em IEEE/ASME Transactions on Mechatronics}, vol.~19, no.~3, pp.~827--839, 2014.

\bibitem{haufle2012clutched}
D.~F. H{\"a}ufle, M.~Taylor, S.~Schmitt, and H.~Geyer, ``A clutched parallel elastic actuator concept: {Towards} energy efficient powered legs in prosthetics and robotics,'' in {\em 2012 4th IEEE RAS \& EMBS International Conference on Biomedical Robotics and Biomechatronics (BioRob)}, pp.~1614--1619, IEEE, 2012.

\bibitem{ossadnik2022BSA}
D.~Ossadnik, M.~C. Yildirim, F.~Wu, A.~Swikir, H.~T.~M. Kussaba, S.~Abdolshah, and S.~Haddadin, ``{BSA} - {Bi-Stiffness Actuation} for optimally exploiting intrinsic compliance and inertial coupling effects in elastic joint robots,'' in {\em Proc. 2022 IEEE/RSJ Int. Conf. on Intell. Robots and Syst. (IROS)}, pp.~3536--3543, 2022.

\bibitem{Pozo23_RAL}
E.~{Pozo Fortuni\'{c}}, M.~C. Yildirim, D.~Ossadnik, A.~Swikir, S.~Abdolshah, and S.~Haddadin, ``{O}ptimally {C}ontrolling the {T}iming of {E}nergy {T}ransfer in {E}lastic {J}oints: {E}xperimental {V}alidation of the {B}i-{S}tiffness {A}ctuation {C}oncept,'' {\em Robotics and Automation Letters (RA-L)}, 2023.
\newblock Under review, arXiv preprint available: https://doi.org/10.48550/arXiv.2309.07873.

\bibitem{chen2013optimal}
L.~Chen, M.~Garabini, M.~Laffranchi, N.~Kashiri, N.~G. Tsagarakis, A.~Bicchi, and D.~G. Caldwell, ``Optimal control for maximizing velocity of the {C}omp{A}ct™ compliant actuator,'' in {\em 2013 IEEE International Conference on Robotics and Automation}, pp.~516--522, IEEE, 2013.

\bibitem{plooij2016reducing}
M.~Plooij, M.~Wisse, and H.~Vallery, ``Reducing the energy consumption of robots using the bidirectional clutched parallel elastic actuator,'' {\em IEEE Transactions on Robotics}, vol.~32, no.~6, pp.~1512--1523, 2016.

\bibitem{Krimsky}
E.~Krimsky and S.~H. Collins, ``{Optimal Control of an Energy-Recycling Actuator for Mobile Robotics Applications},'' in {\em 2020 IEEE International Conference on Robotics and Automation (ICRA)}, pp.~3559--3565, 2020.

\bibitem{van2007introduction}
A.~J. Van Der~Schaft and H.~Schumacher, {\em An introduction to hybrid dynamical systems}, vol.~251.
\newblock springer, 2007.

\bibitem{stewart2000implicit}
D.~Stewart and J.~C. Trinkle, ``An implicit time-stepping scheme for rigid body dynamics with {C}oulomb friction,'' in {\em Proceedings 2000 ICRA. Millennium Conference. IEEE International Conference on Robotics and Automation. Symposia Proceedings (Cat. No. 00CH37065)}, vol.~1, pp.~162--169, IEEE, 2000.

\bibitem{mombaur2005open}
K.~D. Mombaur, R.~W. Longman, H.~G. Bock, and J.~P. Schl{\"o}der, ``Open-loop stable running,'' {\em Robotica}, vol.~23, no.~1, pp.~21--33, 2005.

\bibitem{Posa2014}
M.~Posa, C.~Cantu, and R.~Tedrake, ``A direct method for trajectory optimization of rigid bodies through contact,'' {\em The International Journal of Robotics Research}, vol.~33, no.~1, pp.~69--81, 2014.

\bibitem{nurkanovic2020limits}
A.~Nurkanovi{\'c}, S.~Albrecht, and M.~Diehl, ``Limits of {MPCC} formulations in direct optimal control with nonsmooth differential equations,'' in {\em 2020 European Control Conference (ECC)}, pp.~2015--2020, IEEE, 2020.

\bibitem{sleiman2019contact}
J.-P. Sleiman, J.~Carius, R.~Grandia, M.~Wermelinger, and M.~Hutter, ``Contact-implicit trajectory optimization for dynamic object manipulation,'' in {\em 2019 IEEE/RSJ International Conference on Intelligent Robots and Systems (IROS)}, pp.~6814--6821, IEEE, 2019.

\bibitem{zhu2015optimal}
F.~Zhu and P.~J. Antsaklis, ``Optimal control of hybrid switched systems: A brief survey,'' {\em Discrete Event Dynamic Systems}, vol.~25, pp.~345--364, 2015.

\bibitem{saccon2014sensitivity}
A.~Saccon, N.~Van De~Wouw, and H.~Nijmeijer, ``Sensitivity analysis of hybrid systems with state jumps with application to trajectory tracking,'' in {\em 53rd IEEE Conference on Decision and Control}, pp.~3065--3070, IEEE, 2014.

\bibitem{Haddadin2012}
S.~Haddadin, F.~Huber, and A.~Albu-Schäffer, ``Optimal control for exploiting the natural dynamics of variable stiffness robots,'' in {\em Proc. 2012 IEEE Int. Conf. on Robot. and Autom.}, pp.~3347--3354, 2012.

\bibitem{Andersson2019}
J.~A.~E. Andersson, J.~Gillis, G.~Horn, J.~B. Rawlings, and M.~Diehl, ``{CasADi} -- {A} software framework for nonlinear optimization and optimal control,'' {\em Math. Programming Comput.}, vol.~11, no.~1, pp.~1--36, 2019.

\bibitem{wachter2006implementation}
A.~W{\"a}chter and L.~T. Biegler, ``On the implementation of an interior-point filter line-search algorithm for large-scale nonlinear programming,'' {\em Math. programming}, vol.~106, no.~1, pp.~25--57, 2006.

\bibitem{heger2021biomechanische}
H.~Heger and V.~Wank, ``{B}iomechanische {G}rundlagen des {W}erfens,'' {\em Sportphysio}, vol.~9, no.~01, pp.~8--16, 2021.

\end{thebibliography}
\end{document}